\documentclass[10pt,twocolumn,letterpaper]{article}

\usepackage{cvpr}
\usepackage{times}
\usepackage{epsfig}
\usepackage{graphicx}
\usepackage{amsmath}
\usepackage{amssymb}
\usepackage{color}
\usepackage{symbols}
\usepackage{multirow}
\usepackage{subfigure} 
\usepackage{enumitem} 
\usepackage{caption}
\usepackage{array}
\usepackage{booktabs}
\usepackage{pifont} 
\usepackage{ctable}

\newcommand{\specialcell}[2][l]{%
      \begin{tabular}[#1]{@{}l@{}}#2\end{tabular}}

\makeatletter
\newcommand\footnoteref[1]{\protected@xdef\@thefnmark{\ref{#1}}\@footnotemark}
\makeatother

\usepackage[pagebackref=true,breaklinks=true,letterpaper=true,colorlinks,bookmarks=false]{hyperref}

\cvprfinalcopy % *** Uncomment this line for the final submission

\def\cvprPaperID{6218} % *** Enter the CVPR Paper ID here

\ifcvprfinal\fi
\begin{document}

%%%%%%%%% TITLE
\title{Fast, Diverse and Accurate Image Captioning Guided By Part-of-Speech}

\author{Aditya Deshpande$^{1}$\thanks{~Denotes equal contribution.}, Jyoti Aneja$^{1}$\footnotemark[1], Liwei Wang$^{2}$, Alexander Schwing$^{1}$ and David Forsyth$^{1}$\\
{\tt\small ardeshp2,janeja2,aschwing,daf@illinois.edu \hspace{1.5cm} liweiwang@tencent.com} \\ 
$^{1}$University of Illinois at Urbana Champaign \hspace{2cm} $^{2}$Tencent AI Lab, Seattle
}

\maketitle
%\thispagestyle{empty}

%%%%%%%%% ABSTRACT
% !TEX root = divcap.tex
\begin{abstract}
%Automatically describing an image is an important capability for virtual 
%assistants. Significant progress has been achieved in recent years on this 
%task of image captioning. 

Image captioning is an ambiguous problem, with many suitable
captions for an image. To address ambiguity, beam search is the de facto method for sampling  
multiple captions.  %to address this ambiguity. 
However, beam search is 
computationally expensive and known to produce generic captions~\cite{Finkel2006,Gimpel13asystematic}. 
To address this concern, some variational auto-encoder (VAE)~\cite{WangVAECaption} 
and generative adversarial net (GAN)~\cite{DaiICCV17,shetty2017speaking} 
based methods have been proposed. Though diverse, GAN and VAE 
are less accurate. In this paper, we first predict a meaningful summary of the image, 
then generate the caption based on that summary. We use part-of-speech as
summaries, since our summary should drive caption generation.
We achieve the trifecta: 
(1) High accuracy for the diverse captions as evaluated by standard 
captioning metrics and user studies; (2) Faster computation of diverse 
captions compared to beam search and diverse beam search~\cite{VijayakumarCSSL18}; and 
(3) High diversity as evaluated by counting novel sentences, 
distinct $n$-grams and mutual overlap (\ie, mBleu-4) scores.
\end{abstract}

%%%%%%%%% BODY TEXT
\section{Introduction}

In this paper we show how to force an image captioning system to generate diverse captions by conditioning  on different high-level summaries of the image.
Our summaries are quantized part-of-speech (POS) tag sequences.  Our system generates captions by (a) predicting different summaries from the image then (b) predicting
captions conditioned on each summary.  This approach leads to captions that are {\it accurate, quick to obtain}, and {\it diverse}.
Our system is accurate, because it is able to steer a number of narrow beam searches to explore the space of caption sequences more efficiently.
It is fast because each beam is narrow.  And the captions are diverse, because depending on the summary (\ie, part-of-speech) the system is forced to produce captions that contain (for example) more or less adjectives. 
This means we can avoid the tendency to produce minimal or generic captions that is common in systems that try to optimize likelihood without awareness of language priors (like part-of-speech).

A large body of literature has focused on developing predictive image captioning techniques, often using recurrent
neural nets (RNN)~\cite{Mao2014DeepCW,show_tell,show_attend_tell,neuraltalk,Panderson}.
More recently~\cite{AnejaConvImgCap17,Wang2018CNNCNNCD}, demonstrate
predictive captioning with accuracy similar to RNNs while using convolutional networks.
An essential feature of captioning is that it is ambiguous -- many captions can describe the same image.
This creates a problem, as image captioning programs trained to maximize some score may do so by producing
strongly non-committal captions.  It also creates an opportunity for research -- how can one produce multiple,
diverse captions that still properly describe the image?  Our method offers a procedure to do so. 

\begin{table}[!t]
\vspace{-.4cm}
\begin{center}
\begin{tabular}{l|ccc}
\toprule
        Method & Fast & Diverse  & Accurate \\
\toprule
        Beam search & $\times$ & $\times$ & \checkmark \\
        Diverse beam search~\cite{VijayakumarCSSL18} & $\times$ & $\times$ & \checkmark \\
        AG-CVAE~\cite{WangVAECaption} & \checkmark & $\checkmark$ & $\times$ \\
        Ours (POS) &  \checkmark &  \checkmark &  \checkmark \\
\toprule
\end{tabular}
\vspace{-.4cm}
        \caption{We show that our part-of-speech (POS) based method achieves the trifecta
                of {\bf high accuracy, fast computation} and {\bf more diversity}. Beam search and
                diverse beam search are slow. They also produce captions with high mutual
                overlap and lower distinct $n$-grams than POS (see mBleu-4, div-1 and
                div-2 in \tabref{tab:diversity}). POS and AG-CVAE are fast, however POS
		does better on captioning metrics in \figref{fig:topk} and is therefore
		more accurate.}
\vspace{-1cm}
\label{tab:summary}
\end{center}
\end{table}

Tractable image captioning involves factoring the sequence model for the caption.  Inference then requires beam search,
which investigates a set of captions determined by local criteria to find the caption with highest posterior probability.
Finding very good captions requires a wide beam, which is slow.  Moreover, beam search
is also known to generate generic captions that lack diversity~\cite{Finkel2006,Gimpel13asystematic}.
Variational auto-encoder (VAE)~\cite{WangVAECaption} and generative
adversarial net (GAN)~\cite{DaiICCV17,shetty2017speaking,li2018generating} formulations
outperform beam search on diversity metrics.  VAE and GAN-based methods sample latent vectors from some distribution, then
generate captions conditioned on these samples.  The latent variables have no exposed semantics, and
captions tend not to score as well (on captioning metrics) as those produced by beam search (\eg, Tab.\ 1 of~\cite{shetty2017speaking}).

This paper offers an alternative.  First predict a meaningful summary of the image, then generate the caption based on that summary.  For
this to work, the summary needs to be able to drive language generation (for the caption generator), and must be predictable. We find
quantized part of speech tag sequences to be very effective summaries.  These sequences can clearly drive language generation (\eg, forcing a
captioner to produce adjectives in particular locations).  More surprisingly, one can predict quantized tag sequences from images rather well,
likely because such sequences do summarize the main action of the image.  For example, compare {\it determiner-noun-verb} with {\it determiner-adjective-noun-verb-adjective-noun}. 
In the first case, something appears in the image, in the second, a subject with a noteworthy attribute is doing something to an object with a noteworthy attribute. Consequently, the 
two images appear quite different.

%Our approach offers the inference procedure an approximate
%global view of the sentence being generated at each step. It can then generate a caption that
%adheres roughly to this global view. Sampling different POS tag
%sequences permits to generate multiple captions.  This is roughly equivalent to running several small beams in quite different locations in
%caption space, and our results confirm that our method is very efficient at finding good captions (beam search requires huge, very slow beams
%to beat our approach). Our procedure produces diverse  sentences because sentences that adhere to different tag sequences have meaningful
%diversity.

{\bf Contributions:} We show that image
          captioning with POS tag sequences is fast, diverse and accurate (\tabref{tab:summary}). Our POS
          methods sample captions faster and with more diversity than techniques based on beam
	  search and its variant diverse beam search~\cite{VijayakumarCSSL18} (\tabref{tab:diversity}). Our
          diverse captions are more accurate than their counterparts produced by GANs~\cite{shetty2017speaking} 
          (\tabref{tab:gan}) and VAEs~\cite{WangVAECaption} (\tabref{tab:consensus},
          \figref{fig:topk}).

% !TEX root = divcap.tex
\section{Related Work}
\label{sec:relwork} 

In the following, we first review works that generate a single (or best-1) caption before discussing diverse image captioning methods which produce $k$ different 
(or a set of best-$k$) captions. % ($k = 1, 2, 3, \cdots$). 

\subsection{Image Captioning}

Most image captioning approaches~\cite{neuraltalk,show_tell,show_attend_tell}
use a convolutional neural net pre-trained on classification~\cite{Simonyan14c} to
represent image features. Image features are fed into a recurrent net (often 
based on long-short-term-memory (LSTM) units) to model the language word-by-word. These networks are trained 
with maximum likelihood. 
To obtain high performance on standard image captioning metrics, Yao \etal~\cite{TYao} use a network trained on COCO-attributes in addition 
to image features.
Anderson \etal~\cite{Panderson} develop an attention-based network architecture. 
Aneja \etal~\cite{AnejaConvImgCap17} change the language decoder from an 
LSTM-net to a convoluational network and show that they obtain more diversity.
Similarly, Wang \etal~\cite{Wang2018CNNCNNCD} also use a convolutional language
decoder. Since diversity is of interest to us, we use the convolutional 
language decoder similar to~\cite{AnejaConvImgCap17,Wang2018CNNCNNCD}. We leave incorporation of techniques such as 
 attribute vectors specific to the COCO dataset, and a sophisticated 
attention mechanism from~\cite{TYao,Panderson} for further performance gains 
to future work. 

Apart from exploring different network architectures, some prior works focus on using 
different training losses. Reinforcement learning has been used 
in~\cite{luo2018discriminability,rennie2017self,liu2016improved}, to 
directly train for  non-differentiable evaluation metrics such as BLEU, 
CIDEr and SPICE. In this paper, we use maximum likelihood training for 
our methods and baselines to ensure a fair comparison. 
Training our POS captioning network in a 
reinforcement learning setup can be investigated as part of future work. 

Notable advances have been made in conditioning image captioning on semantic priors of objects by 
using object detectors~\cite{Lu2018Neural,WangObjectCounts}. This conditioning  
is only limited to the objects (or nouns) in the caption and ignores the remainder, 
while our POS approach achieves coordination for the entire sentence.

\subsection{Diverse Image Captioning}

Four main techniques have been proposed to generate multiple captions 
and rank them to obtain a set of best-$k$ captions. %($k = 1, 2, 3, \cdots$).

\noindent
{\bf Beam search.} Beam search is the classical method to sample multiple 
solutions given sequence models for neural machine translation and image 
captioning. We compare  
to beam search on the same base captioning network as POS, but without 
part-of-speech conditioning. We find that though beam search is 
accurate, it is slow (\tabref{tab:consensus}) and lacks diversity (\tabref{tab:diversity}). 
Our base captioning network uses a convolutional neural net (CNN)~\cite{AnejaConvImgCap17} and 
is equivalent to the standard LSTM based captioning network of 
Karpathy \etal~\cite{neuraltalk} in terms of accuracy. 

\noindent
{\bf Diverse beam search.} Vijayakumar \etal~\cite{VijayakumarCSSL18} 
augment  beam search  with an additional diversity function to 
generate diverse outputs. They propose a hamming diversity function 
that penalizes expanding a beam with the same word used in an earlier 
beam. In our results, we compare to this diverse beam search (Div-BS).
Note, beam search and diverse beam search are local search procedures 
which explore the output captioning space word-by-word. While, POS tag 
sequences act as global probes that permit to sample captions 
in many different parts of the captioning space. 

\noindent
{\bf GAN.} More recent work on diverse image captioning 
focuses on using GANs. % and VAEs.  We review work on GANs first and then VAE. 
Adversarial training has been employed 
by~\cite{DaiICCV17,li2018generating,shetty2017speaking} to generate 
diverse captions.~\cite{DaiICCV17,li2018generating} train a conditional 
GAN for diverse caption generation.~\cite{DaiICCV17} uses a trainable loss 
which differentiates human annotations from generated captions. Ranking 
based techniques, which attempt to score human annotated captions higher than 
generated ones, are demonstrated in~\cite{li2018generating}. 
Shetty \etal~\cite{shetty2017speaking} use adversarial training in combination 
with an approximate Gumbel sampler to match the generated captions to the 
human annotations. 

Generally, GAN based methods improve on diversity, but suffer on accuracy. For 
example, in Tab.~1 of~\cite{shetty2017speaking}, METEOR and SPICE scores drop 
drastically compared to an LSTM baseline. In \tabref{tab:gan}, 
we compare GAN~\cite{shetty2017speaking} and our POS-based method which is more accurate.

\noindent
{\bf VAE.} Wang \etal~\cite{WangVAECaption} propose to generate diverse captions 
using a conditional variational auto-encoder with an additive Gaussian latent space 
(AG-CVAE) instead of a GAN. The diversity obtained with their approach is due to  
sampling from the learned latent space. They demonstrate improvements in 
accuracy over the conventional LSTM baseline. Due to the computational complexity of beam search they used fewer  
beams for the LSTM baseline compared to the number of captions sampled from the VAE, \ie, they ensured equal computational time. 
We compare to AG-CVAE~\cite{WangVAECaption} and show that we obtain higher 
best-1 caption accuracy (\tabref{tab:consensus}) and our best-$k^\text{th}$ 
caption accuracy ($k = 1$ to $10$) outperforms AG-CVAE (\figref{fig:topk}). 
Note, best-$k$ scores in \tabref{tab:consensus} and \figref{fig:topk} denote the 
score of the $k^\text{th}$ ranked caption given the same number of sampled captions 
(20 or 100) for all methods. For fairness, we use the same ranking procedure 
(\ie, consensus re-ranking proposed by~\cite{Devlin2015ExploringNN} and used 
in~\cite{WangVAECaption}) to rank the sampled captions for all methods.

\section{Background}
\label{sec:background}

\noindent
{\bf Problem Setup and Notation.} The goal of diverse captioning is to generate 
$k$ sequences $y^{1}, y^{2}, \ldots, y^{k}$, given an image. For readability
we drop the super-script and focus on a single sequence $y$. The methods we 
discuss and develop will sample many such sequences $y$ and rank them to obtain the best-$k$
-- $y^{1}, y^{2}, \ldots, y^{k}$. A single caption 
$y = (y_1, \ldots, y_N)$ consists of a sequence of words $y_i$, $i\in\{1, \ldots, N\}$ which accurately describe 
the given image $I$.  For each caption $y$, the words $y_i, i\in\{1,\ldots, N\}$ are 
obtained from a fixed vocabulary $\cY$, \ie, $y_i\in\cY$. Additionally, we 
assume availability of a part-of-speech (POS) tagger for the sentence $y$. 
More specifically, the POS tagger 
provides a tag sequence $t = (t_1, \ldots, t_N)$ for a given sentence, 
where $t_i\in\cT$ is the POS tag for word $y_i$. The set $\cT$ encompasses 
12 universal POS tags -- {\it verb (VERB), noun (NOUN), pronoun (PRON)}, 
\etc\footnote{\label{nltk} See \url{https://www.nltk.org/book/ch05.html} for POS tag and automatic POS tagger details}

To train our models we use a dataset $\cD = \{(I, y, t)\}$ which contains tuples $(I,y,t)$ 
composed of an image $I$, a sentence $y$, and the corresponding POS tag sequence $t$. 
Since it is not feasible to annotate the $\sim.5$M captions of MSCOCO with POS tags, 
 we use an automatic part-of-speech tagger.\footnoteref{nltk}

\begin{figure}[!t]
\vspace{-.3cm}
\centering
\includegraphics[width=.45\textwidth]{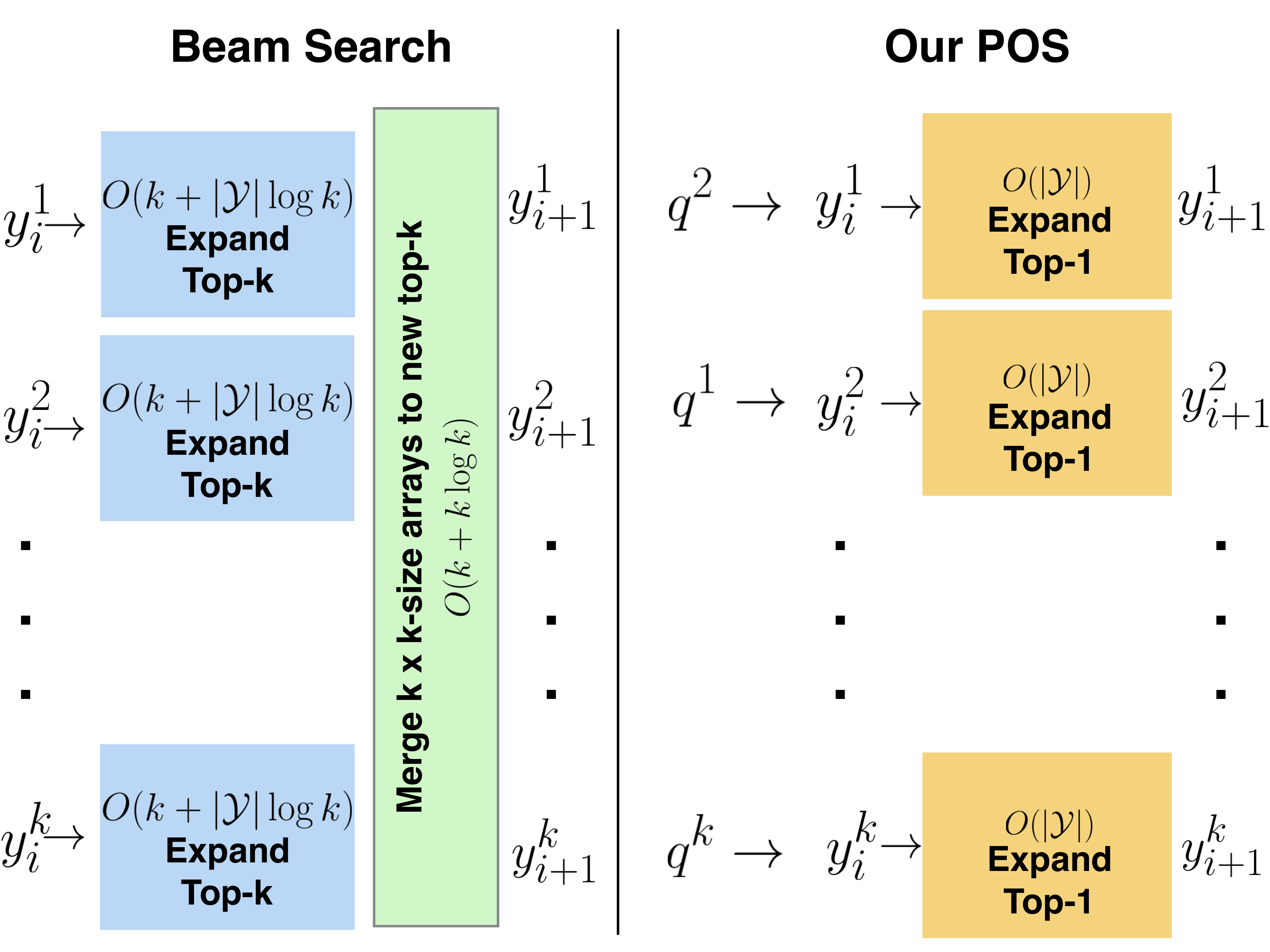}
\vspace{-.4cm}
\caption{Illustration of beam search and POS-sampling to expand the best-$k$ captions 
$(y_i^1, y_i^2, \ldots y_i^k)$ from word position $i$ to $i+1$. See \secref{sec:background}
for  notation and other details.} 
\vspace{-.4cm}
\label{fig:complexity}
\end{figure}

\noindent
{\bf Classical Image Captioning.} Classical techniques factor the joint probabilistic model $p_{\theta}(y|I)$ over all words into a product of conditionals. They learn model parameters $\theta^\ast$ by maximizing the likelihood over the training set $\cD$, \ie,
\begin{equation}
\label{eq:mle}
%\begin{split}
%\theta^\ast \!\!=\!\! \arg
\max\limits_{\theta} \!\!\!\!\sum_{(I, y) \in \cD}\!\!\!\! \log p_{\theta}(y | I), 
\text{~~where~~} 
p_{\theta}(y|I) \!\!=\!\! \prod\limits_{i=1}^{N} p(y_{i} | y_{<i}, I).
%\end{split}
\end{equation}
The factorization of the joint probability distribution enforces a temporal ordering of words. Hence, word $y_i$ at the $i^\text{th}$ time-step  
(or word position) depends only on all previous words $y_{<i}$. This probability model is represented using 
a recurrent neural network or a feed-forward network with temporal (or masked) convolutions. 
Particularly the latter, \ie, temporal convolutions, have been used recently for different vision and language tasks 
in place of classical recurrent neural nets, \eg,~\cite{AnejaConvImgCap17,S2S,ConvEmpBai}. 

\begin{figure*}[!t]
\centering
\vspace{-.3cm}
\includegraphics[width=.85\textwidth]{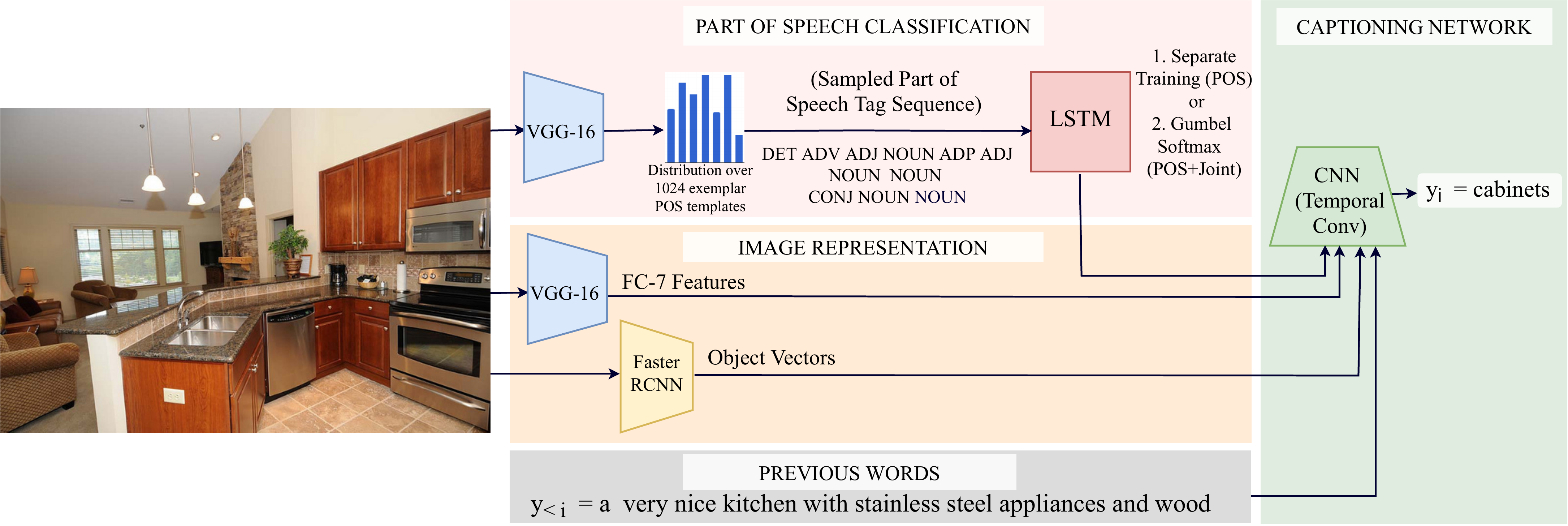}
\vspace{-.3cm}
\caption{An illustration of our POS captioning method on a test image. For the image representation, fc7 features are extracted from VGG-16 and embedded into 512 dimensional vectors. For object vectors, we use the 80 dimensional class vector from faster rcnn~\cite{ren2015faster} (same as~\cite{WangVAECaption}). For part-of-speech classification, we use VGG-16 with two linear layers and a 1024-way softmax. Then, we encode sampled 
POS via an LSTM-net to a 512 dimensional vector. Our captioning network 
uses temporal convolutions and operates on image representation, part-of-speech vector, object vector and previous words in the caption ($y_{<i}$) to produce the next word ($y_i$). The network is trained for 20 epochs using the ADAM optimizer~\cite{ADAM} (initial learning rate of $5e^{-5}$ and a decay factor of .1 after 15 epochs). The part of speech classification step can be trained 
separately (POS) or jointly using a gumbel softmax (POS+Joint). Note, image representation is same for our method and baselines.} 
\vspace{-.3cm}
\label{fig:divcap_model}
\end{figure*}

During training, we learn the optimal parameters $\theta^{*}$. 
Then for test image $I$, conditional word-wise 
posterior probabilities $p_{\theta^{*}}(y_i|y_{<i}, I)$ are generated sequentially 
from $i = 1$ to $N$. Given these posteriors, beam search is applicable and forms our baseline. 
\figref{fig:complexity} illustrates beam search with a  beam width of $k$ 
from word position $y_i$ to $y_{i+1}$. Here, beam search maintains best-$k$ 
(incomplete) captions ordered by likelihood. It expands the best-$k$ captions at 
every word greedily from start to end of the sentence.  

More specifically, for beam search from word position $i$, we first generate posteriors 
$p_{\theta^{*}}^j(y_{i+1} | y^j_{<(i+1)}, I)$ based on the current top-$k$ list containing $y^j_{<(i+1)}$, $j \in\{ 1, \ldots, k\}$. 
We then obtain new top-$k$ captions by  expanding each of the $k$ entries $y^j_{<(i+1)}$ in the list using the 
computed posterior $p_{\theta^{*}}^j(y_{i+1} | y^j_{<(i+1)}, I)$. We call this `expand top-$k$.'
The time complexity for a single expand top-$k$ operation is identical to obtaining the sorted 
top-$k$ values from an array of size $|\cY|$.\footnote{\url{https://www.geeksforgeeks.org/k-largestor-smallest-elements-in-an-array/}} 
The time complexity of all expand top-$k$ operations is $O(k^2 + |\cY| k \log k)$. 

We merge all the %$k$ 
expanded top-$k$ captions to
the final top-$k$ captions using the log sum of the posterior probability at word
position $i+1$. We call this operation merge. The merge operation has a complexity of  
 $O(k + k \log k)$, which is identical to merging $k$ sorted arrays.\footnote{\url{https://www.geeksforgeeks.org/merge-k-sorted-arrays/}}
 In \secref{sec:poscap}, we show that our inference with POS has better time complexity.

\begin{table*}[!t]
\vspace{-.3cm}
\centering
\resizebox{.95\textwidth}{!}{%
\begin{tabular}{l|c|llllllll|l|cc}
	  Method & \multirow{2}{*}{\specialcell{Beam size \\ or \#samples}}& \multicolumn{8}{c|}{Best-1 Oracle Accuracy} & \multirow{2}{*}{\specialcell{ Speed \\ (s/img)}} & Speed & Accuracy\\ 
	& & B4 & B3 & B2 & B1 & C &  R & M & S & & & \\
\toprule
	Beam search & \multirow{5}{*}{20}  & 0.489$^\checkmark$ &  0.626$^\checkmark$ & 0.752$^\checkmark$ & 0.875$^\checkmark$ & 1.595$^\checkmark$ & 0.698$^\checkmark$ & 0.402$^\checkmark$ & 0.284$^\checkmark$ & 3.74$^ \times$ & $\times$ & \checkmark \\ 
	Div-BS~\cite{VijayakumarCSSL18} & & 0.383$^\times$ & 0.538$^\times$ & 0.687$^\times$ & 0.837 & 1.405 & 0.653  & 0.357  & 0.269 &  3.42 & $\times$ & $\times$ \\ 
	AG-CVAE~\cite{WangVAECaption} & & 0.471 & 0.573 & 0.698 & 0.834$^\times$ & 1.308$^\times$ & 0.638$^\times$ & 0.309$^\times$ & 0.244$^\times$ & - & - & $\times$ \\
	POS & & 0.449 & 0.593 & 0.737 & 0.874 & 1.468 & 0.678 & 0.365 & 0.277 & 0.21 & \checkmark & \checkmark \\
	POS+Joint & & 0.431 & 0.581 & 0.721 & 0.865 & 1.448 & 0.670 & 0.357 & 0.271 & 0.20$^\checkmark$ & \checkmark & \checkmark \\
\toprule
	Beam Search & \multirow{5}{*}{100}   & 0.641$^\checkmark$ & 0.742$^\checkmark$ & 0.835$^\checkmark$ & 0.931$^\checkmark$ & 1.904$^\checkmark$ & 0.772$^\checkmark$ & 0.482$^\checkmark$ & 0.332$^\checkmark$ & 20.33 & $\times$ & \checkmark \\ 
	Div-BS~\cite{VijayakumarCSSL18}  & & 0.402$^\times$  & 0.555$^\times$ & 0.698$^\times$ & 0.846$^\times$  &  1.448$^\times$ & 0.666$^\times$ & 0.372  & 0.290 &  19.05 & $\times$ & $\times$ \\ 
	AG-CVAE~\cite{WangVAECaption} & & 0.557 & 0.654 & 0.767 & 0.883 & 1.517 & 0.690 & 0.345$^\times$ & 0.277$^\times$ & - & - & $\times$ \\
	POS & & 0.578 & 0.689 & 0.802 & 0.921 & 1.710 & 0.739 & 0.423 & 0.322 & 1.29  & \checkmark & \checkmark \\
	POS+Joint & & 0.550 & 0.672 & 0.787 & 0.909 & 1.661 & 0.725 & 0.409 &  0.311 & 1.27$^\checkmark$ & \checkmark & \checkmark \\
\end{tabular}%
}
\vspace{-.3cm}
\caption{{\bf Best-1 accuracy by oracle re-ranking}. Our POS methods are faster at sampling than beam search and they also 
generate a higher scoring best-1 caption than AG-CVAE~\cite{WangVAECaption} and Div-BS~\cite{VijayakumarCSSL18}. 
Beam search obtains the best scores, however it is slow. From all sampled captions 
(\#samples = $20$ or $100$), we use oracle to pick the best-1 caption for every 
metric. This gives an estimate of the upper bound on captioning accuracy for each 
method.  We use standard captioning metrics, BLEU (B1-B4)~\cite{BLEU}, CIDEr (C)~\cite{Rama}, 
ROUGE (R)~\cite{ROUGE}, METEOR (M)~\cite{METEOR} and SPICE (S)~\cite{SPICE}. Note, 
\checkmark indicates good performance on the metric for the corresponding 
column and $\times$ indicates bad performance.}
\vspace{-.3cm}
\label{tab:oracle}
\end{table*}

\section{Image Captioning with Part-of-Speech}
\label{sec:poscap}

In our approach for image captioning, we introduce a POS tag sequence $t$, 
to condition the recurrent model given in \equref{eq:mle}.  More formally, we use the distribution  
\begin{equation}
p_{\theta}(y|t, I) = \prod\limits_{i=1}^{N}  p_{\theta}(y_{i} | t, y_{<i}, I).
\end{equation}
Following classical techniques, we train our POS-conditioned approach by maximizing the likelihood (similar to  \equref{eq:mle}), \ie, we want to find the parameters %$\theta^\ast$ via
\begin{equation}
\theta^\ast = \argmax\limits_{\theta} \sum\limits_{(I, t, y) \in \cD} \log p_{\theta}(y | t, I).
\label{eq:mlepos}
\end{equation} 
Importantly, note that we use the entire POS tag sequence in the conditional above, because it allows 
global control over the entire sentence structure. 

Training involves learning the  parameters $\theta^\ast$ for our
conditional captioning model (\equref{eq:mlepos}). During test time, 
 conditioning on POS tags provides a mechanism for diverse image captioning, \ie,  
given a test image $I$, we obtain $k$ diverse captions by sampling $k$  
POS tag sequences $t^1, t^2, \ldots, t^k$. 
Note that every  sequence is a tuple of 
POS tags, \ie,  $t^i = (t^i_1, t^i_2, \ldots)$, $i\in\{1, \ldots, k\}$. 

Since a large number of possible POS tag sequences exists, in \secref{sec:posclassify}, we %show that we can sample quantized POS 
discuss how we obtain quantized POS 
tag sequences  $q^1, q^2, \ldots, q^k$ given the input 
image. These quantized sequences approximate the actual POS tag sequences 
$t^1, t^2, \ldots, t^k$. %In our notation, $q$ is the quantized tag 
%sequence corresponding to $t$. 

Concretely, during inference we sample $k$ quantized POS tag 
sequences  given the image. This is shown as the part-of-speech 
classification step in \figref{fig:divcap_model}. Then, we encode 
each sampled POS tag sequence $q$ using an LSTM model. The encoded POS tag
sequence, along with object vector, image features (fc7 of VGG-16) 
and previous words ($y_{<i}$) forms the input to the temporal 
convolutions-based captioning network. This captioning network 
implements our posterior probability $p_{\theta}(y_i | y_{<i}, q, I)$, which is used to 
 predict the next word $y_i^\ast = \argmax_{ y_i} p_{\theta}( y_i | y_{<i}, q, I)$. 

\noindent
{\bf Fast inference with POS.} For every sampled tag sequence $q^j, j \in \{1, 2, \cdots, k \}$  
(\ie quantization of tag sequence $t^j$), we maximize the learned probabilistic model, 
\ie, $y_{i}^j = \argmax_{y} p_{\theta^{*}}(y_{i} | y_{<i}, q^j, I)$ greedily. As just discussed, we 
 simply use the maximum probability word at every word 
position. \figref{fig:complexity} compares this computationally much more effective method, which has a time complexity of $O(k|\cY|)$, to the breadth first
approach employed by beam search. 

Note that POS-based 
sampling requires only a single $\max$-operation at every step during inference (our effective
beam size is 1), making it %theoretically 
faster than beam search with wide beams. It is also faster than diverse beam search (with group size parameter 
set to 1 as in our results) which performs the $k$ `expand top-$k$' operations %in beam search 
sequentially using an augmented diversity function. 

\subsection{Image to Part-of-Speech Classification}
\label{sec:posclassify}
Because our model conditions sentence probabilities on a POS tag sequence, 
we need to compute it before performing inference. %inference requires supplying a sequence. 
Several ways exist to obtain the 
POS tag sequence. \Eg, choosing a POS tag sequence by hand,  sampling from a 
distribution of POS tag sequences seen in the dataset $\cD$, or predicting POS tag sequences 
conditioned on the observed image $I$. The first one is not scalable. The second approach 
of sampling from $\cD$ without considering the provided image is easy, but generates inaccurate captions. We found  
the third approach to yield most accurate results. While this  seems like an odd task at first, our experiments suggest 
very strongly that image based prediction of POS tag sequences works rather well. Indeed, intuitively, inferring  a POS tag sequence from an image is similar 
to predicting a situation template~\cite{yatskar2016} -- one must predict a 
rough template sketching what is worth to be said about an image. 

To capture multi-modality, we use a classification model to compute our POS predictions for a given image $I$. However, we find that there are $> 210K$ POS tag sequences in our training 
dataset $\cD$ of $|\cD| > 500K$ captions.
To maintain efficiency, we therefore quantize the space of POS tag sequences to 1024 exemplars as discussed subsequently. 
%This requires  to vector quantize the set of all observed 
%POS tag sequence. We quantize to 1024 (exemplar) sequences. 

\noindent
{\bf Quantizing POS tag sequences.}
 We perform a hamming distance 
based k-medoids clustering to obtain $1024$-cluster centers. We use concatenated 1-hot encodings (of POS tags) to
encode the POS tag sequence. We observe our clusters to be tight, \ie,  
more than $75\%$ of the clusters have an average hamming distance less than 3. 
We use the cluster medoids as the quantized POS tag sequences for our classifier. 
Given an input tag sequence $t$ we represent it using its nearest neighbor in quantized space, which we denote 
 by $q = \mathcal{Q}(t)$. Note, in our notation the quantization function 
$\mathcal{Q}(t)$, reduces $t$ to its quantized tag sequence $q$. 

Our image to part-of-speech classifier (shown in \figref{fig:divcap_model}) 
learns to predict over quantized POS sequence space by maximizing the likelihood, 
$p_{\phi}(q | I)$. Formally, we look for its optimal parameters $\phi^\ast$ via
\begin{equation}
	\label{eq:imgpos}
%	\begin{split}
	\phi^\ast = \argmax\limits_{\phi} \sum\limits_{(I, t) \in \cD} \log p_{\phi}(q | I), %\\ 
%	\text{where}\quad 
%	\end{split}
\end{equation}
where $\log p_{\phi}(q | I) = \sum\limits_{i = 1}^{1024} \delta[q^{i} = \mathcal{Q}(t)] \log p_{\phi}(q^{i} | I)$.

%Now, with this knowledge of quantization at hand, note that we replace $t$ with its corresponding 
%EC-POS template $\mathcal{Q}(t)$ in \equref{eq:mlepos} above.

\subsection{Separate vs.\ Joint Training}
\label{sec:sepjoint}

Training involves learning the parameters $\theta$ of the captioning network
 (\equref{eq:mlepos}) and the parameters $\phi$ of the POS classification 
network  (\equref{eq:imgpos}). We can trivially train these two networks
separately and we call this method {\bf POS}. 

%\noindent
%{\bf Joint Training.} 
We also experiment with joint training by sampling from 
the predicted POS posterior $p_{\phi}(t | I)$  using a Gumbel soft-max~\cite{gumbel} before subsequently using its output in the captioning network. 
%This makes joint part-of-speech classification end-to-end trainable with the
%captioning network. 
%This permits 
Inconsistencies between sampled POS sequence and corresponding 
caption $y$ will introduce noise since the ground-truth caption $y$
may be incompatible with the sampled sequence $q$. Therefore, during every training iteration, we sample $50$ POS tag 
sequences from the Gumbel soft-max and only pick the one $q$ with the best 
alignment to POS tagging of caption $y$. We refer 
to this form of joint training via {\bf POS+Joint}. In \secref{sec:top1} and 
\secref{sec:topk}, we show that POS+Joint (\ie, jointly learning $\theta$ 
and $\phi$) is useful and produces more accurate captions.

% !TEX root = divcap.tex
\section{Results}
\label{sec:res} 
In the following, we compare our developed approach for diverse captioning with POS tags to competing baselines for diverse captioning. We first provide information about the dataset, the baselines and the evaluation metrics before presenting our results. 

\begin{table*}[!t]
\centering
\resizebox{.9\textwidth}{!}{%
\begin{tabular}{l|c|llllllll|l|cc}
	Method & \multirow{2}{*}{\specialcell{Beam size \\ or \#samples}}  & \multicolumn{8}{c|}{Best-1 Consensus Re-ranking Accuracy} & \multirow{2}{*}{\specialcell{ Speed \\ (s/img)}} & Speed & Accuracy \\ 
	& & B4 & B3 & B2 & B1 & C &  R & M & S & & & \\
\toprule

	\specialcell{Beam search \\ (w. Likelihood)} & \multirow{5}{*}{20} & 0.305 & 0.402$^\times$ & 0.538$^\times$ & 0.709$^\times$ & 0.947$^\times$ & 0.523 & 0.248 & 0.175 & 3.19 & $\times$ & $\times$ \\
	Beam search & & 0.319 & 0.423 & 0.564 & 0.733 & 1.018 & 0.537$^\checkmark$ & 0.255 & 0.185 & 7.41 & $\times$ & \checkmark \\
	Div-BS~\cite{VijayakumarCSSL18}  & & 0.320$^\checkmark$ & 0.424$^\checkmark$ & 0.562 &  0.729 &   1.032$^\checkmark$ &  0.536 &  0.255$^\checkmark$ &  0.184 &  7.60$^\times$ & $\times$ & \checkmark \\
	AG-CVAE~\cite{WangVAECaption} & & 0.299$^\times$ & 0.402$^\times$ & 0.544 & 0.716 & 0.963 & 0.518$^\times$ & 0.237$^\times$ & 0.173$^\times$ & - & - & $\times$ \\
	POS  & & 0.306 & 0.419 & 0.570$^\checkmark$ & 0.744$^\checkmark$ & 1.014 & 0.531 & 0.252 & 0.188$^\checkmark$ & 1.13$^\checkmark$ & \checkmark & \checkmark \\
	POS+Joint  & & 0.305 & 0.415 & 0.563 & 0.737 & 1.020 & 0.531 & 0.251 & 0.185 & 1.13$^\checkmark$ & \checkmark & \checkmark \\
\toprule
  
	\specialcell{Beam search \\ (w. Likelihood)} & \multirow{5}{*}{100} & 0.300$^\times$ & 0.397$^\times$ & 0.532$^\times$ & 0.703$^\times$ & 0.937$^\times$ & 0.519$^\times$ & 0.246 & 0.174$^\times$ & 18.24 & $\times$ & $\times$ \\
	Beam search  & & 0.317 & 0.419 & 0.558 & 0.729 & 1.020 & 0.532 & 0.253 & 0.186 & 40.39$^\times$ & $\times$ & \checkmark \\
	Div-BS~\cite{VijayakumarCSSL18} & & 0.325$^\checkmark$ & 0.430$^\checkmark$ & 0.569$^\checkmark$ &  0.734 & 1.034 & 0.538$^\checkmark$ &  0.255$^\checkmark$ &  0.187 &  39.71 & $\times$ & \checkmark \\
	AG-CVAE~\cite{WangVAECaption} & & 0.311 & 0.417 & 0.559 & 0.732 & 1.001 & 0.528 & 0.245$^\times$ & 0.179 & - & - & $\times$ \\
	POS & & 0.311 & 0.421 & 0.567 & 0.737 & 1.036 & 0.530 & 0.253 & 0.188$^\checkmark$ & 7.54 & \checkmark & \checkmark \\
	POS+Joint & & 0.316 & 0.425 & 0.569$^\checkmark$ & 0.739$^\checkmark$ & 1.045$^\checkmark$ & 0.532 & 0.255$^\checkmark$ & 0.188$^\checkmark$ & 7.32 & \checkmark & \checkmark \\
\end{tabular}%
}
\vspace{-.3cm}
\caption{{\bf Best-1 accuracy by consensus re-ranking}. Our POS methods obtain higher scores on 
captioning metrics than AG-CVAE~\cite{WangVAECaption}. This demonstrates our POS natural 
language prior is more useful than the abstract latent vector used by VAE-based methods. POS methods 
obtain comparable accuracy to Beam Search and Div-BS~\cite{VijayakumarCSSL18}, and they are more
computationally efficient at sampling (\ie, high speed). Note, we 
also outperform the standard beam search that uses likelihood based ranking. For these results, consensus 
re-ranking~\cite{Devlin2015ExploringNN} is used to pick the best-1 caption from all sampled captions 
(unless `w. Likelihood' is specified). For fair comparison, each method uses the same 80-dimensional 
object vector  from faster rccn~\cite{fasterRCNN} and the same image features/parameters for 
consensus re-ranking. The captioning metrics are the same as in \tabref{tab:oracle}. Note, 
\checkmark indicates good performance on the metric for the corresponding 
column and $\times$ indicates bad performance.}
\vspace{-.3cm}
\label{tab:consensus}
\end{table*}

\begin{table}
\vspace{-.3cm}
\begin{center}
\setlength{\tabcolsep}{2pt}
\begin{tabular}{l|c|cc}
%\toprule
Method & \#samples & Meteor & Spice\\
\toprule
Beam Search (with VGG-16)  & 5 & .247 & .175 \\
GAN (with Resnet-152)  & 5 & .236 & .166 \\
POS+Joint (with VGG-16)~\cite{shetty2017speaking} & 5 & {\bf .247} & {\bf .180} \\
%\toprule
\end{tabular}
\vspace{-.3cm}
\caption{{\bf Comparison to GAN-based method.} To compare to GAN, we train 
our POS+Joint on another split of MSCOCO by Karpathy \etal~\cite{neuraltalk}. Our POS+Joint 
method samples more accurate best-1 captions than the GAN method. POS+Joint also obtains better 
SPICE score on this split compared to beam search. Our accuracy may improve with 
the use of Resnet-152 features. For fair comparison, we use the same 80-dimensional 
object vectors from faster rcnn~\cite{fasterRCNN} and rank the generated captions with 
likelihood for all methods.}
\vspace{-1cm}
\label{tab:gan}
\end{center}
\end{table}

\noindent
{\bf Dataset.} We use the~\textbf{MS COCO} dataset~\cite{MSCOCO} for our experiments. 
For the train/val/test splits we follow: (1) M-RNN~\cite{Mao2014DeepCW} using 118,287 
 images for training, 4,000 images for validation, and 1,000 images for testing; and 
(2) Karpathy \etal~\cite{neuraltalk} using 113,287  images for training, 5,000 
images for validation and 5,000 images for testing. The latter split is used to compare to GAN-based results in  
\tabref{tab:gan}. 
 
\noindent
{\bf Methods.} In the results, we denote our approach by {\bf POS}, and 
our approach with joint training by {\bf POS+Joint} (see \secref{sec:sepjoint} for the differences). 
We compare to the additive Gaussian conditional VAE-based diverse captioning 
method of Wang \etal~\cite{WangVAECaption},  denoted by {\bf AG-CVAE}. Our captioning 
network is based on~\cite{AnejaConvImgCap17}.  For a fair comparison 
to beam search we also compare to convolutional captioning~\cite{AnejaConvImgCap17}  with beam search. 
This is referred to as {\bf beam search}.  We compare to diverse beam search denoted denoted
by {\bf Div-BS}. The abbreviation {\bf GAN} is used to denote the GAN-based 
method in~\cite{shetty2017speaking}.  

\noindent
{\bf Evaluation criteria.} We compare all methods using four
criteria -- accuracy, diversity, speed, human perception:
\begin{itemize}[itemsep=-1ex,topsep=0pt,partopsep=0pt]
\item {\bf Accuracy.} In \secref{sec:top1} (Best-1 Accuracy) we compare the accuracy using the standard 
image captioning task of generating a single caption. Subsequently,  in 
\secref{sec:topk} (Best-$k^\text{th}$ Accuracy), we 
assess the accuracy of $k$ captions on different image captioning metrics. 
\item {\bf Diversity.} We evaluate the performance of each method on different diversity 
metrics in \secref{sec:diversity}. 
\item {\bf Speed.} In addition to accuracy, in \secref{sec:speed}, we also measure the computational 
efficiency of each method for sampling multiple captions. % in \tabref{tab:oracle} and \tabref{tab:consensus}.
\item {\bf Human perception.} We do a user study in \secref{sec:ustudy}. 
\end{itemize} 

\begin{figure*}[!t]
\vspace{-.5cm}
  \centering
  \subfigure[Best-10 CIDEr from 20 samples]{\includegraphics[width=.24\textwidth]{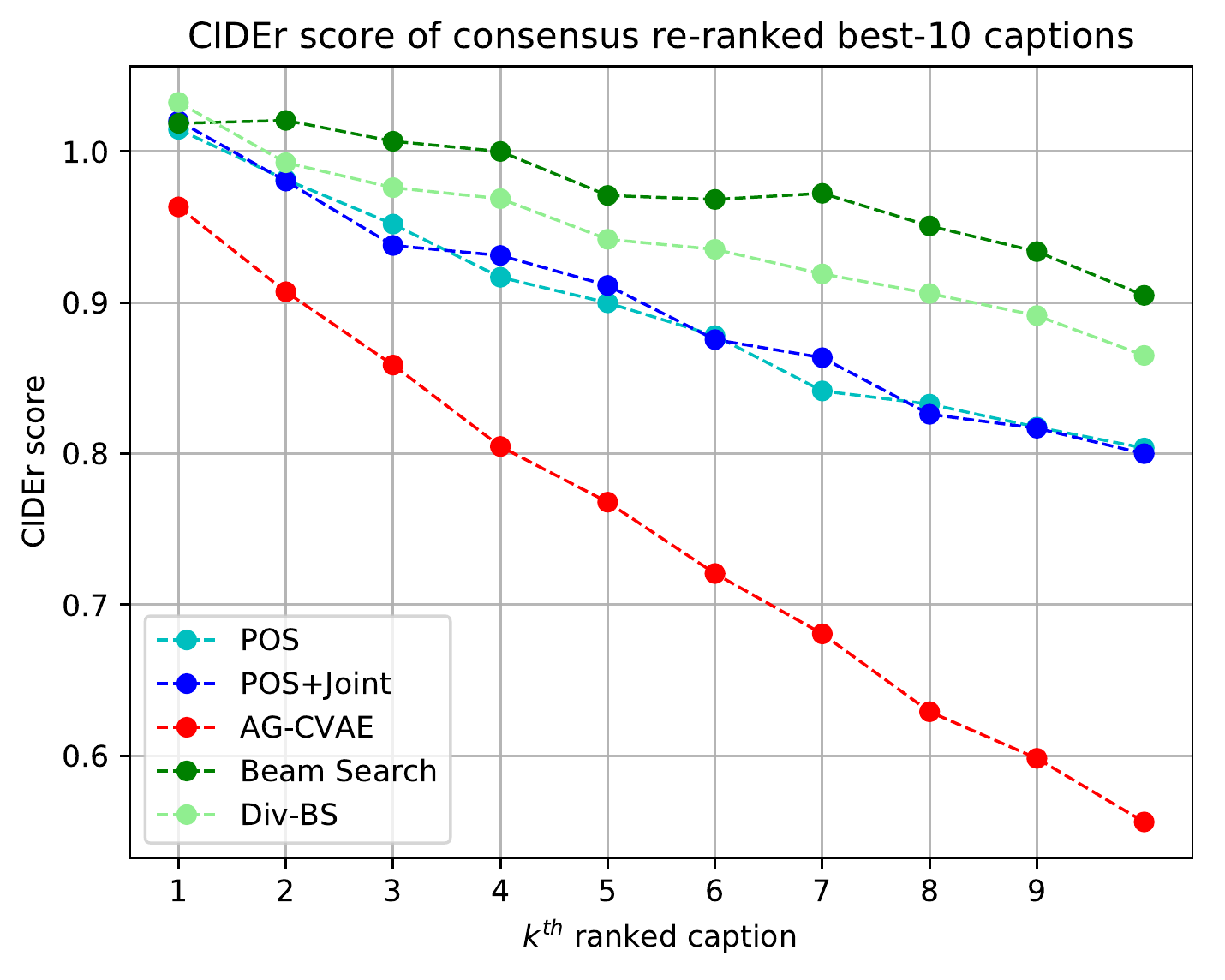}}
  \subfigure[Best-10 SPICE from 20 samples]{\includegraphics[width=.24\textwidth]{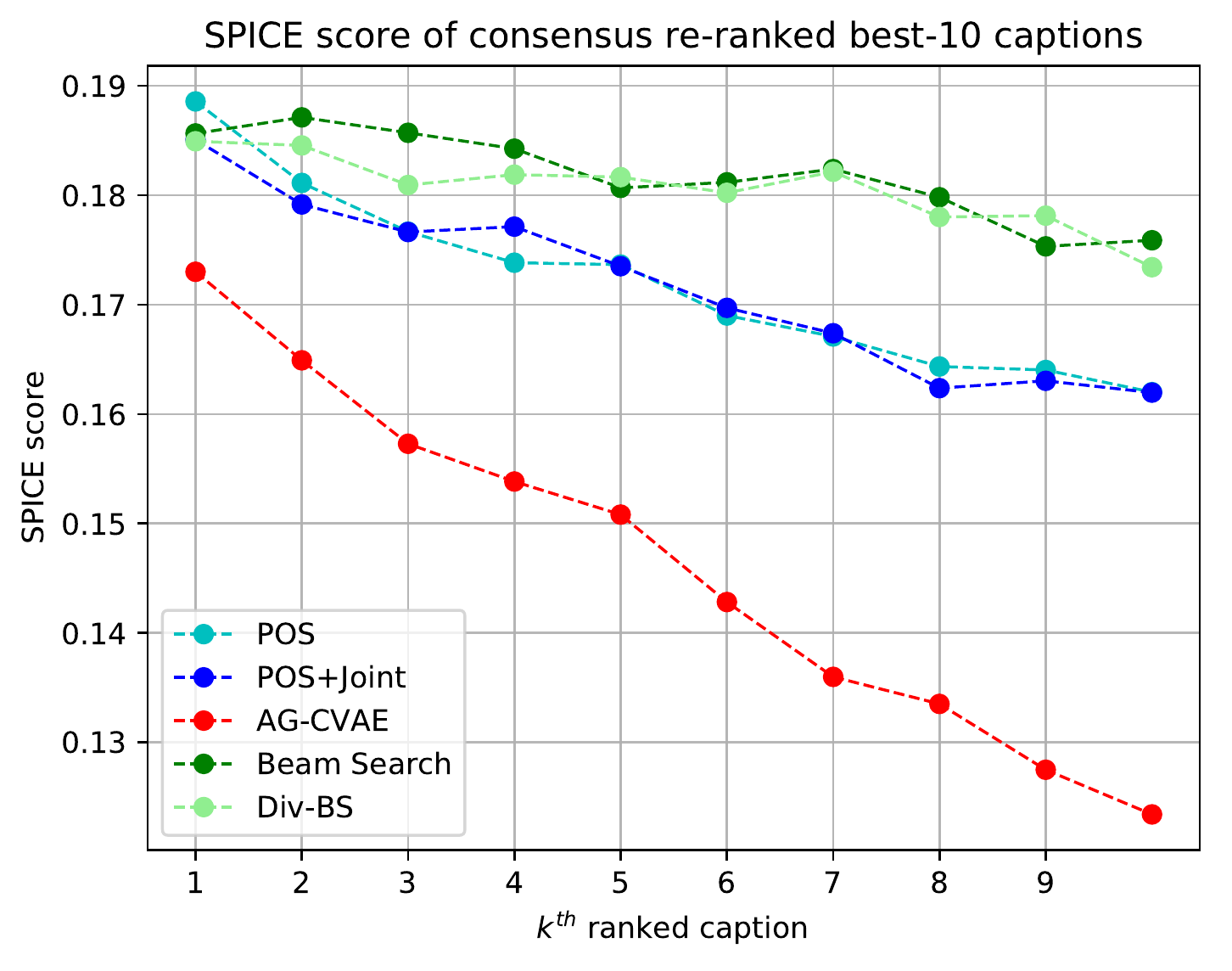}}
  \subfigure[Best-10 CIDEr from 100 samples]{\includegraphics[width=.24\textwidth]{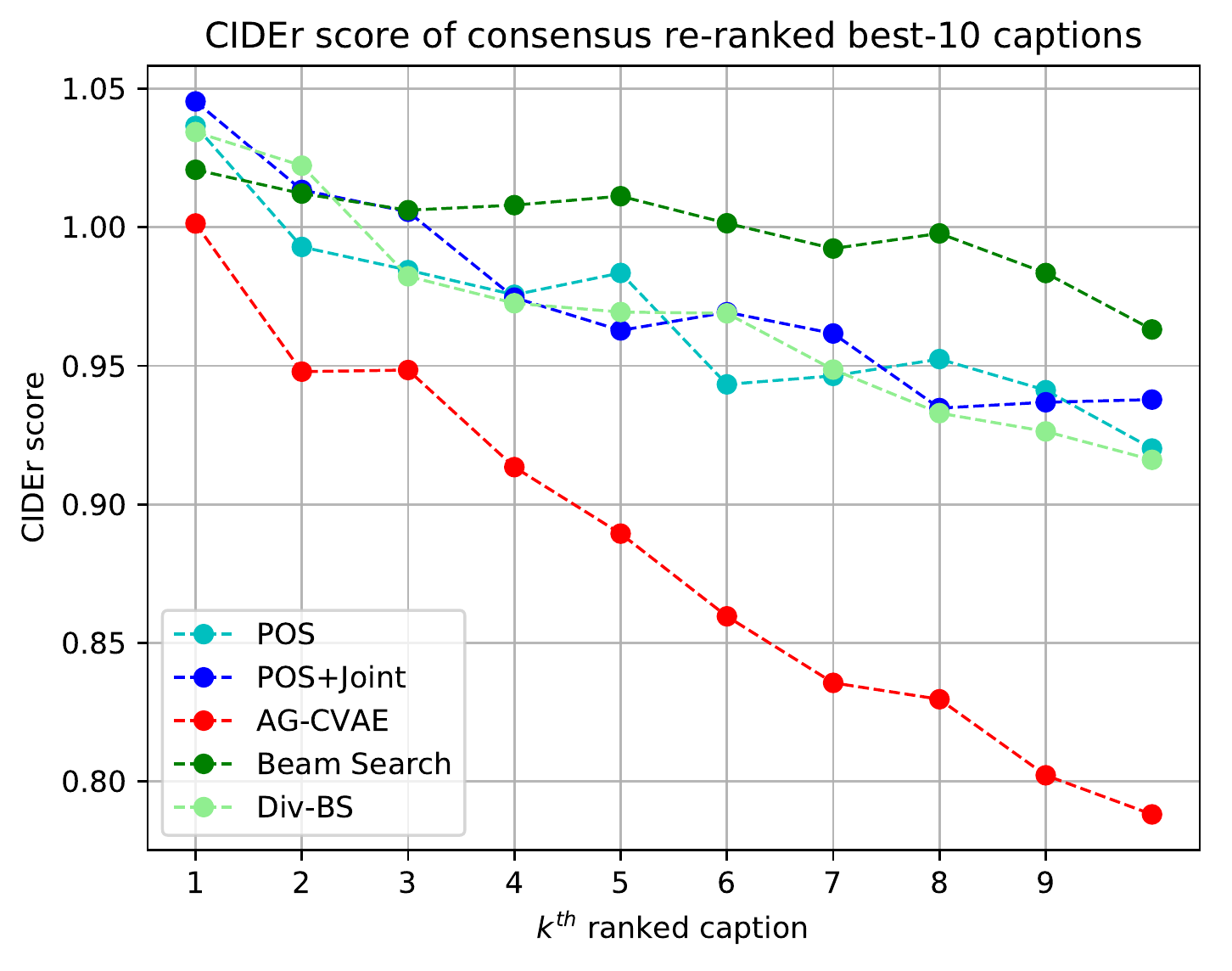}}
  \subfigure[Best-10 SPICE from 100 samples]{\includegraphics[width=.24\textwidth]{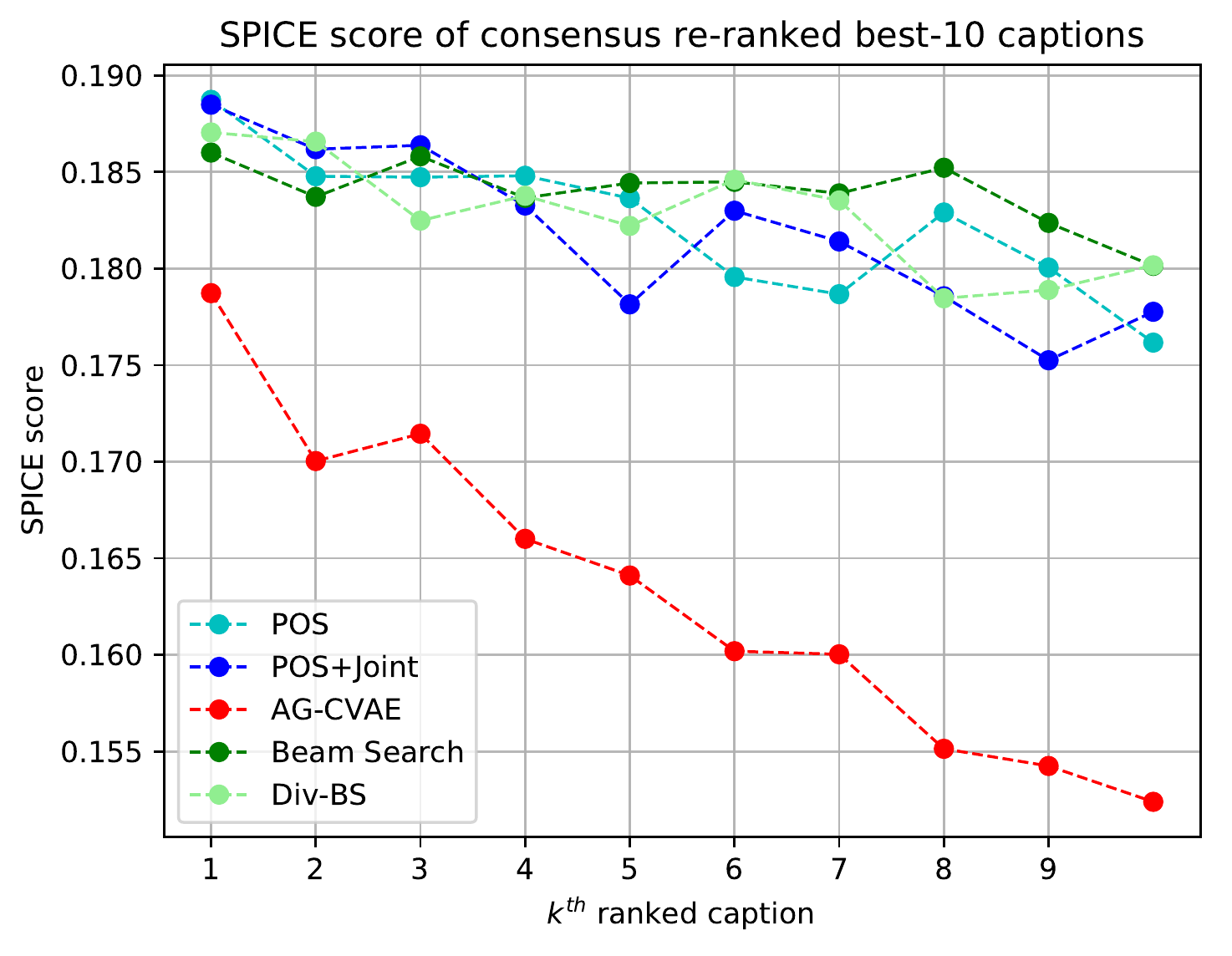}}
\vspace{-.5cm}
	\caption{{\bf Best-10 CIDEr and SPICE accuracy.} Our POS and POS+Joint achieve best-$k$ accuracy comparable
	   to Beam Search and Div-BS~\cite{VijayakumarCSSL18} with faster computation time. We outperform the best-$k$ scores
	   of AG-CVAE~\cite{WangVAECaption}, demonstrating part-of-speech conditioning is better than abstract latent 
	   variables of a VAE. Note, this figure is best viewed in high-resolution.} 
\vspace{-.3cm}
  \label{fig:topk}
\end{figure*}

\begin{table*}[!t]
%\vspace{-.3cm}
\begin{center}
\setlength{\tabcolsep}{2pt}
\resizebox{.8\textwidth}{!}{%
\begin{tabular}{l|c|c|c|c|cc|c}
	Method & Beam size & Distinct& \# Novel sentences & mBleu-4 & \multicolumn{2}{c|}{$n$-gram Diversity (Best-5)} & Overall Diversity\\
	& or \#samples & Captions & (Best-5)  & (Best-5)  & Div-1 & Div-2 & \\
\toprule
	Beam search & \multirow{5}{*}{20} &  {\bf 100\%} &  2317 & 0.777 & 0.21 & 0.29 & $\times$ \\
	Div-BS~\cite{VijayakumarCSSL18} & &  {\bf 100\%} &  3106 & 0.813 & 0.20 & 0.26 & $\times$\\
	AG-CVAE~\cite{WangVAECaption}  & & 69.8\% & 3189 & 0.666  & {\bf 0.24} & 0.34 & \checkmark \\
	POS & & 96.3\% & 3394 & {\bf 0.639} & {\bf 0.24} & {\bf 0.35} & \checkmark\\
	POS+Joint & & 77.9\% & {\bf 3409} &  0.662 & 0.23 & 0.33 & \checkmark \\
\toprule
	Beam search & \multirow{5}{*}{100} & {\bf 100\%} & 2299 & 0.781 & 0.21 & 0.28 & $\times$ \\
	Div-BS~\cite{VijayakumarCSSL18} & & {\bf 100\%} & 3421  & 0.824 & 0.20 & 0.25 & $\times$ \\
	AG-CVAE~\cite{WangVAECaption} & & 47.4\% & 3069 & 0.706 & {\bf 0.23} & 0.32 & \checkmark \\
	POS & & 91.5\% & {\bf 3446} & {\bf 0.673} & {\bf 0.23} & {\bf 0.33} & \checkmark \\
	POS+Joint & & 58.1\% & 3427 & 0.703 & 0.22 & 0.31 & \checkmark \\
\specialrule{.2em}{.1em}{.1em}
	Human & 5 & 99.8\% & - &  0.510 &  0.34 &  0.48 \\
\end{tabular}%
}
\vspace{-.3cm}
\caption{{\bf Diversity statistics.} For each method, we report the number of novel sentences (\ie, 
sentences not seen in the training set) out of at most best-5 
sentences after consensus re-ranking. Though Beam Search showed high 
accuracy in \tabref{tab:oracle}, \ref{tab:consensus} and \figref{fig:topk}, 
here, we see that it produces less number of novel sentences than our POS 
methods. Therefore, beam search is more prone to regurgitating training data. Low 
mBleu-4 indicates lower 4-gram overlap between generated captions and more diversity
in generated captions. POS has the lowest mBleu-4 and therefore high diversity in generated 
captions. For details on other metrics see \secref{sec:diversity}.}
\vspace{-.6cm}
\label{tab:diversity}
\end{center}
\end{table*}

\subsection
{\bf Best-1 Accuracy} 
\label{sec:top1}

We use two ranking methods -- oracle and consensus re-ranking --
on the set of generated captions  and pick the best-1 caption. 
Our results for oracle re-ranking in \tabref{tab:oracle} and for consensus re-ranking in \tabref{tab:consensus} show that, 
beam search and diverse beam search are accurate however slow. POS
is both fast and accurate. POS outperforms the accuracy of AG-CVAE. 

\noindent
{\bf Oracle re-ranking.} The reference captions of the test set are used and 
the generated caption with the maximum score for each metric is chosen as best-1
(as also used in~\cite{WangVAECaption}). This metric permits to assess the best caption  for each metric and the score  provides an upper bound on the achievable 
best-1 accuracy. Higher oracle scores are also indicative of the method 
being a good search method in the space of captions. Results in 
\tabref{tab:oracle} show that beam search obtains the best oracle scores. 
However, it is painfully slow ($\sim20$s per image to sample 100 captions). 
POS, POS+Joint obtain higher accuracy than AG-CVAE and comparable 
accuracy to beam search with faster runtime. 

\noindent
{\bf Consensus re-ranking scores.} In a practical test setting, 
reference captions of the test set won't be available to rank
the best $k$ captions and obtain best-1. Therefore, in consensus 
re-ranking, the reference captions of training images similar to the test 
image are retrieved. The generated captions are ranked via the CIDEr 
score computed with respect to the retrieved reference set~\cite{Devlin2015ExploringNN}. 

We use the same image features~\cite{WangLL17} and parameters for consensus 
re-ranking as~\cite{WangVAECaption}. \tabref{tab:consensus} shows that our methods 
POS and POS+Joint  outperform the AG-CVAE baseline on all metrics. Moreover, our methods 
are faster than beam search and diverse beam search. They produce higher 
CIDEr, Bleu-1,2, METEOR and SPICE scores. Other scores are comparable and 
differ in the $3^\text{rd}$ decimal. Note, our POS+Joint achieves better scores 
than POS, especially for 100 samples. This demonstrates that joint 
training is useful. 

We also train our POS+Joint method on the train/test split of Karpathy \etal~\cite{neuraltalk}
used by the GAN method~\cite{shetty2017speaking}. In \tabref{tab:gan}, we show that we 
obtain higher METEOR and SPICE scores than those reported in~\cite{shetty2017speaking}. 

\begin{table}
\begin{center}
\setlength{\tabcolsep}{2pt}
\begin{tabular}{l|c|c}
%\toprule
	Baseline Method & POS Wins & Baseline Method Wins  \\
\toprule
	Beam search & {\bf 57.7\%} & 42.2\% \\
	Diverse beam search~\cite{VijayakumarCSSL18} & 45.3\% & {\bf  54.6\%} \\
	AG-CVAE~\cite{WangVAECaption} & {\bf 64.8\%} & 35.1\% 
%\toprule
\vspace{-.4cm}
\end{tabular}
	\caption{We show the user captions sampled from best-$k$ (same $k^\text{th}$ ranked, $k=1$ to 5) for 
	baseline methods and our POS. The user is allowed to pick the caption that best describes the
	image. Note, user is not aware of the method that generated the caption. Here, we observe that
	our POS method outperforms Beam search and AG-CVAE on our user study. Our user study has
	$123$ participants with on average $23.3$ caption pairs annotated by each user.}
\vspace{-1cm}
\label{tab:ustudy}
\end{center}
\end{table}

\begin{figure*}[!t]
\vspace{-.5cm}
\centering
  \subfigure[Qualitative Comparison]{\includegraphics[width=.6\textwidth,trim={0cm 0cm 0cm 0cm},clip]{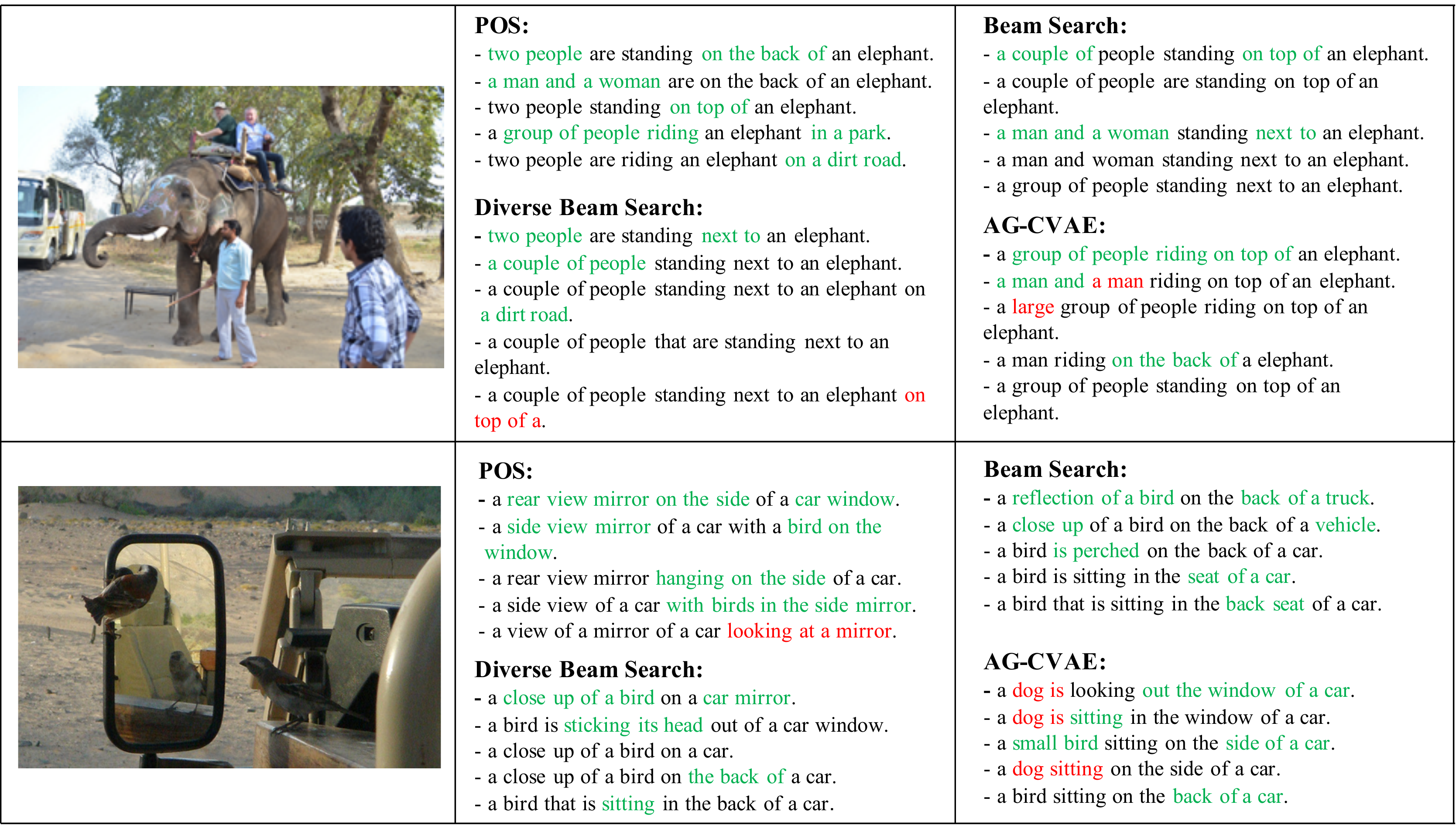}}
  \subfigure[Diversity (or Overlap) Comparison]{\includegraphics[width=.3\textwidth]{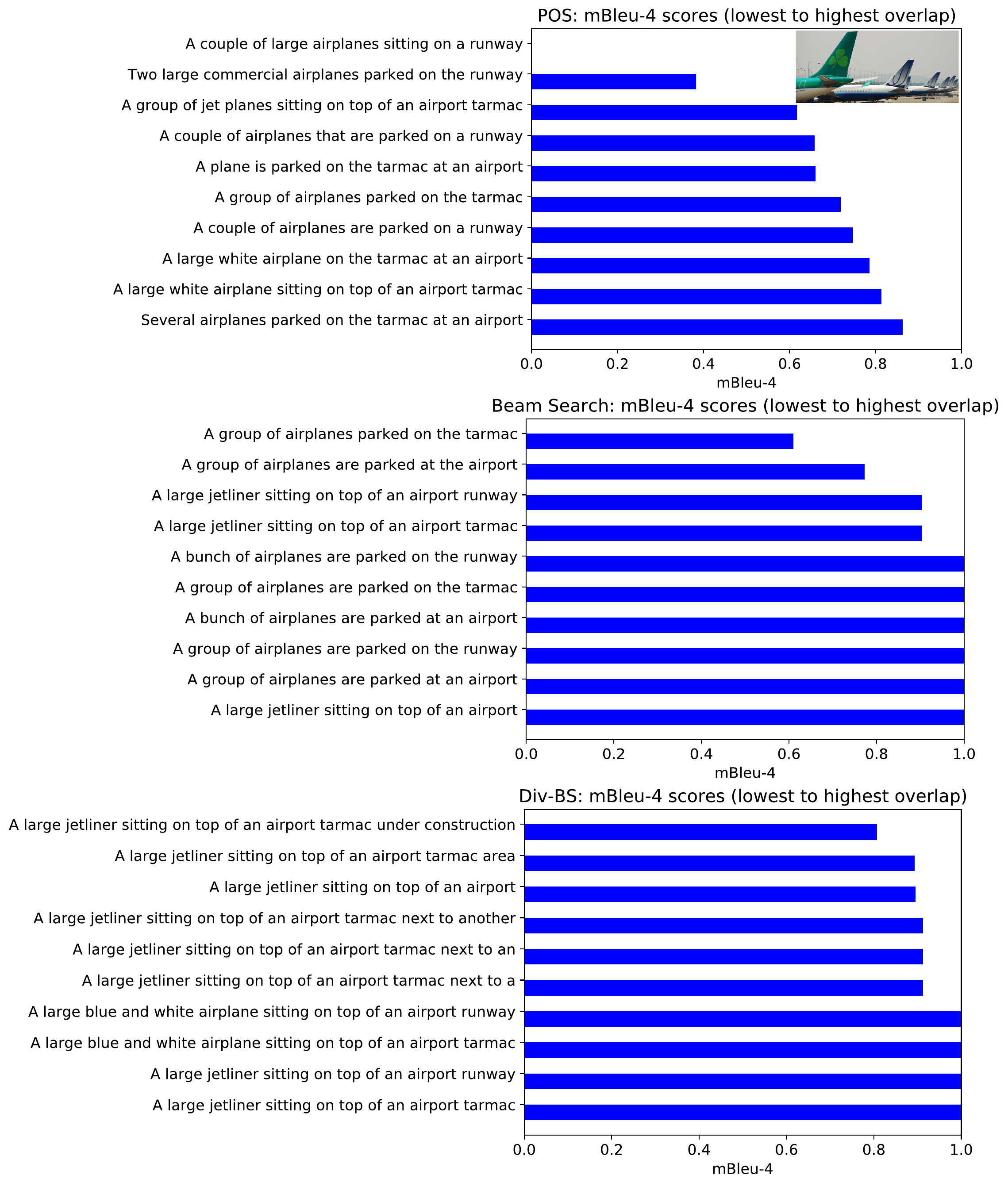}}
\vspace{-.4cm}
\caption{In figure on left, notice POS captions contain things like rear/side view mirror, dirt road, 
	the quantifier `two' which is less common in other methods. The inaccuracies are highlighted in red 
	and the novel parts in green. In figure on right, we compare the diversity (or overlap) of captions.
	The mBleu-4 score measures 4-gram overlap between one generated caption and the rest. Lower is better, 
	\eg, 0 means caption has no 4-gram overlap to other sentences. POS is better than BS and Div-BS
	in the plots above (lower mBleu-4 scores). Note, ground-truth 5 captions all have 0 overlap to each 
	other for this example. On our 1000 image test set with 10 captions generated per image, POS generates 
	$10.94\%$ sentences with 0 overlap; in contrast Div-BS generates $1.02\%$ and Beam Search $2.4\%$. 
	Figure best viewed in high-resolution.}
\vspace{-.3cm}
\label{fig:qual}
\vspace{-.3cm}
\end{figure*}

\subsection{Best-$k^\text{th}$ Accuracy}
\label{sec:topk}

Our captioning method can be conditioned on different part-of-speech tags to generate
diverse captions. For diverse image captioning, in addition to best-1 accuracy, best-$k^\text{th}$ accuracy 
should also be measured. Best-$k^\text{th}$ accuracy is the score of the $k^\text{th}$ ranked caption, 
therefore it is lower than the best-1 score. All $k$ generated captions should be accurate 
and therefore it is desirable to have high best-$k^\text{th}$ scores. This 
metric has not been reported previously~\cite{shetty2017speaking,WangVAECaption}. 

In \figref{fig:topk}, we compare best-$k^\text{th}$ ($k = 1$ to $10$) scores for all methods. 
Note, the accuracy of AG-CVAE drops drastically on both CIDEr and Spice, while our POS 
methods maintain accuracy comparable to beam search. This proves that our POS image summaries 
are better at sampling accurate captions than the abstract latent variables of a VAE. 

%{\color{blue}
%I'm missing a mention of \tabref{tab:gan} here to ensure that the tables are mentioned in consecutive order in the narrative
%A: done in best-1 acc above 
%}

\subsection{Evaluation of Diversity}
\label{sec:diversity}
 
In \tabref{tab:diversity} we compare methods on diversity metrics.

\noindent
(1) {\bf Uniqueness.} The number of unique sentences generated after sampling. 
Beam search and diverse beam search always sample a unique sentence. Note, our POS also 
samples a high number of unique sentences 19.26 (96.3\%) out of 20, 91.55 
out of 100. The uniqueness reduces for joint training. 
This is because, generation of a caption while training POS+Joint 
is based on a noisy POS tag sequence sampled from the Gumbel softmax. Therefore, 
the caption may not be compatible with this noisy POS tag sequence which leads to an 
overly smooth latent representation for the POS tag. Therefore, different POS 
tags may produce the same latent code and hence the same caption.

\noindent
(2) {\bf Novel sentences.} We measure the number of novel sentences (not seen in train), 
and find that our POS-based methods produce more novel sentences than  
all other methods. Beam search produces the least number of novel 
sentences. 

\noindent
(3) {\bf Mutual overlap.} We also measure the mutual overlap between generated captions. This
is done by taking one caption out of $k$ generated captions and evaluating
the average Bleu-4 with respect to all other $k-1$ captions. Lower value
indicates higher diversity. POS is the most diverse. Note, the average score
 is computed by picking every caption \vs the remaining 
$k-1$ captions.

\noindent
(4) {\bf n-gram diversity (div-$n$).} We measure the ratio of distinct $n$-grams per caption to the
total number of words generated per image. POS outperforms other methods.

\subsection{Speed}
\label{sec:speed}

In \figref{fig:complexity} we showed that our POS based methods have better time 
complexity than beam search and diverse beam search. The time complexity of our POS-based approach is the same
as sampling from a VAE or GAN, provided the max probability word is chosen at each
word position (as we do). The empirical results in \tabref{tab:oracle} and \tabref{tab:consensus}
show that POS methods are $5 \times$ faster than beam search methods.

\subsection{User Study}
\label{sec:ustudy}

\figref{fig:qual} compares the captions generated by different methods and in \tabref{tab:ustudy},
we provide the results of a user study. A user is shown two captions sampled from two different methods. The user is asked to pick the more appropriate image caption. 
\tabref{tab:ustudy} summarizes our results. We observe POS outperforms AG-CVAE and Beam search. 

%\begin{table}[!t]
%\begin{center}
%\setlength{\tabcolsep}{2pt}
%\resizebox{.45\textwidth}{!}{%
%\begin{tabular}{|l|cccccc|}
%\hline
%	Method & Color & Object & Relation & Attribute & Cardinality & Size \\
%\hline
%\hline
%	Beam search  &  {\bf 0.090} & 0.342 & 0.048 & {\bf 0.086} & 0.022 & 0.033 \\
%	Div-BS~\cite{VijayakumarCSSL18} & 0.083 & 0.344 & 0.047 & 0.084 & {\bf 0.061} & 0.041 \\ 
%	AG-CVAE~\cite{WangVAECaption}  & 0.065 & 0.324 & 0.047 & 0.064 & 0.026 & {\bf 0.058} \\  
%	POS & 0.075 & {\bf 0.351} & {\bf 0.050} & 0.078 & 0.044 & 0.047 \\
%	POS+Joint & 0.076 & 0.350 & 0.049 & 0.077 & 0.044 & 0.049 \\
%\hline
%\end{tabular}%
%}
%\caption{}
%\label{tab:spice}
%\end{center}
%\end{table}

% !TEX root = divcap.tex
\vspace{-0.2cm}
\section{Conclusion} 
\vspace{-0.2cm}
The developed diverse image captioning approach conditions on part-of-speech. 
It obtains higher accuracy (best-1 and best-10) than GAN and VAE-based 
methods and is computationally more efficient than the classical beam 
search. It performs better on different diversity metrics compared 
to other methods. 

%End-to-end training with a learned ranker instead 
%of non-parameteric consensus re-ranking, is a promising direction 
%for future work. 

{\small
\noindent\textbf{Acknowledgments:}  Supported by NSF 
Grant No.\ 1718221 and ONR MURI Award N00014-16-1-2007.}

\clearpage
{\small
\bibliographystyle{ieee}
\bibliography{egbib}
}

\end{document}